\pdfoutput=1
\PassOptionsToPackage{dvipsnames}{xcolor}
\documentclass[11pt]{article}
\usepackage{acl2023}

\usepackage{times}
\usepackage{latexsym}

\usepackage{marvosym}

\usepackage[T1]{fontenc}

\usepackage[utf8]{inputenc}

\usepackage{microtype}

\usepackage{inconsolata}
\usepackage{microtype}

\usepackage{algorithm,algcompatible,amsmath}
\usepackage{graphicx}
\usepackage{url}
\usepackage{multirow}
\usepackage{comment}
\usepackage{amssymb}
\usepackage{pifont}
\usepackage{mathtools}
\usepackage{xcolor}
\usepackage{enumitem}

\usepackage{textcomp}
\usepackage{array}
\usepackage{booktabs}
\usepackage{multicol}
\usepackage{flushend}
\usepackage{amsfonts}
\usepackage{amsmath}
\usepackage{calc}
\usepackage{amssymb}
\usepackage{diagbox}
\usepackage{algorithm}  
\usepackage{algorithmicx}  
\usepackage{algpseudocode} 
\usepackage{ulem}
\usepackage{color}
\usepackage{array}
\usepackage{subfigure}

\definecolor{light-gray}{gray}{0.95}

\makeatletter
\newcommand{\thickhline}{%
    \noalign {\ifnum 0=`}\fi \hrule height 1pt
    \futurelet \reserved@a \@xhline
}

\newcommand{\squishend}{
\end{list} }
\newcommand*\circled[1]{\kern-2.5em%
  \put(0,4){\color{white}\circle*{18}}\put(0,4){\circle{10}}%
  \put(-3,0){\color{black}\bfseries#1}~~}
\usepackage{tikz} 

\usepackage{enumitem}

\newcommand{\squishlist}{
\begin{list}{$\bullet$}
{ \usecounter{Lcount}
\setlength{\itemsep}{0pt}
\setlength{\parsep}{0pt}
\setlength{\topsep}{0pt}
\setlength{\partopsep}{0pt}
\setlength{\leftmargin}{2em}
\setlength{\labelwidth}{1.5em}
\setlength{\labelsep}{0.5em} } }
\newcommand{\modelname}{\textsl{Jointprop}}
\newcommand{\basemodelname}{\textsl{Beforeprop}}
\newcommand{\woRE}{\textsl{w/o REprop}}
\newcommand{\woNER}{\textsl{w/o NERprop}}

\title{Jointprop: Joint Semi-supervised Learning for Entity and Relation Extraction with Heterogeneous Graph-based Propagation}

\author{Zheng Yandan$^{1,2}$, Hao Anran$^1$ and Luu Anh Tuan$^{1}$ \\ $^1$School of Computer Science and Engineering \\  $^2$ Interdisciplinary Graduate Program-HealthTech \\ Nanyang Technological University, Singapore \\
\texttt{\{yandan002, s190003\}@e.ntu.edu.sg} \\ \texttt{anhtuan.luu@ntu.edu.sg} 
       }

\begin{document}
\maketitle

\setlength{\abovedisplayskip}{4pt}
\setlength{\belowdisplayskip}{4pt}

\begin{abstract}

Semi-supervised learning has been an important approach to address challenges in extracting entities and relations from limited data. However, current semi-supervised works handle the two tasks (i.e., Named Entity Recognition and Relation Extraction) separately and ignore the cross-correlation of entity and relation instances as well as the existence of similar instances across unlabeled data. To alleviate the issues, we propose {\modelname}, a Heterogeneous Graph-based Propagation framework for joint semi-supervised entity and relation extraction, which captures the global structure information between individual tasks and exploits interactions within unlabeled data. 
Specifically, we construct a unified span-based heterogeneous graph from entity and relation candidates and propagate class labels based on confidence scores. We then employ a propagation learning scheme to leverage the affinities between labelled and unlabeled samples.
Experiments on benchmark datasets show that our framework outperforms the state-of-the-art semi-supervised approaches on NER and RE tasks. We show that the joint semi-supervised learning of the two tasks benefits from their codependency and validates the importance of utilizing the shared information between unlabeled data. 

\end{abstract}

%

\section{Introduction}


Named Entity Recognition (NER) and Relation Extraction (RE) are two crucial tasks in Information Extraction. 
Supervised learning schemes have made significant progress in NER and RE research by leveraging rich annotated data (e.g., \citet{lin-etal-2020-joint, yamada-etal-2020-luke, baldini-soares-etal-2019-matching}. However, high-quality data annotation still involves extensive and expensive labor. Moreover, training NER and RE models in various domains and applications require diverse annotated data. Semi-supervised learning approaches (SSL) employ a small amount of annotated data as a source of supervision for learning powerful models at a lower cost.

\begin{figure}[t!]
 \centering
    \subfigure[Before label propagation \label{a}]{ 
    \includegraphics[width=0.5\textwidth]{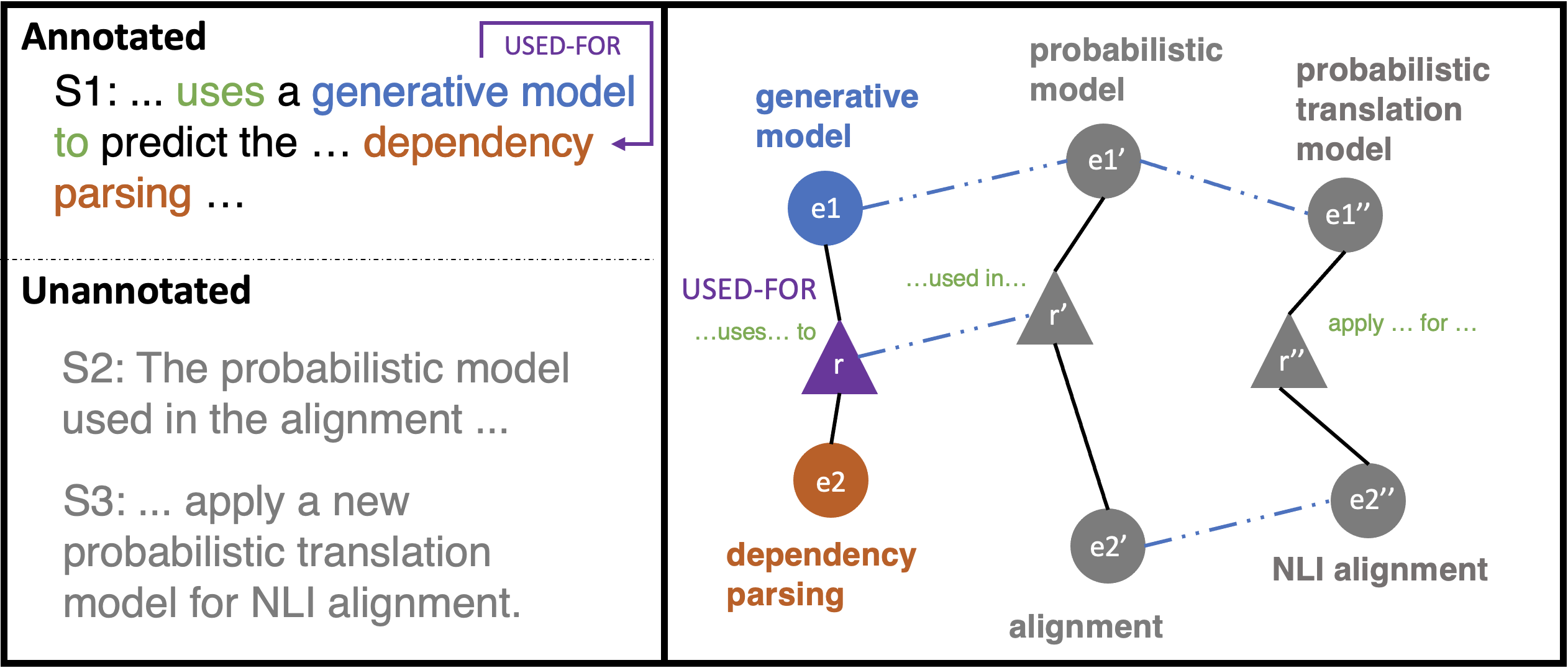}
    \label{fig:figure-1.a}
}
\subfigure[After label propagation\label{b}]{ 
 \includegraphics[width=0.5\textwidth]{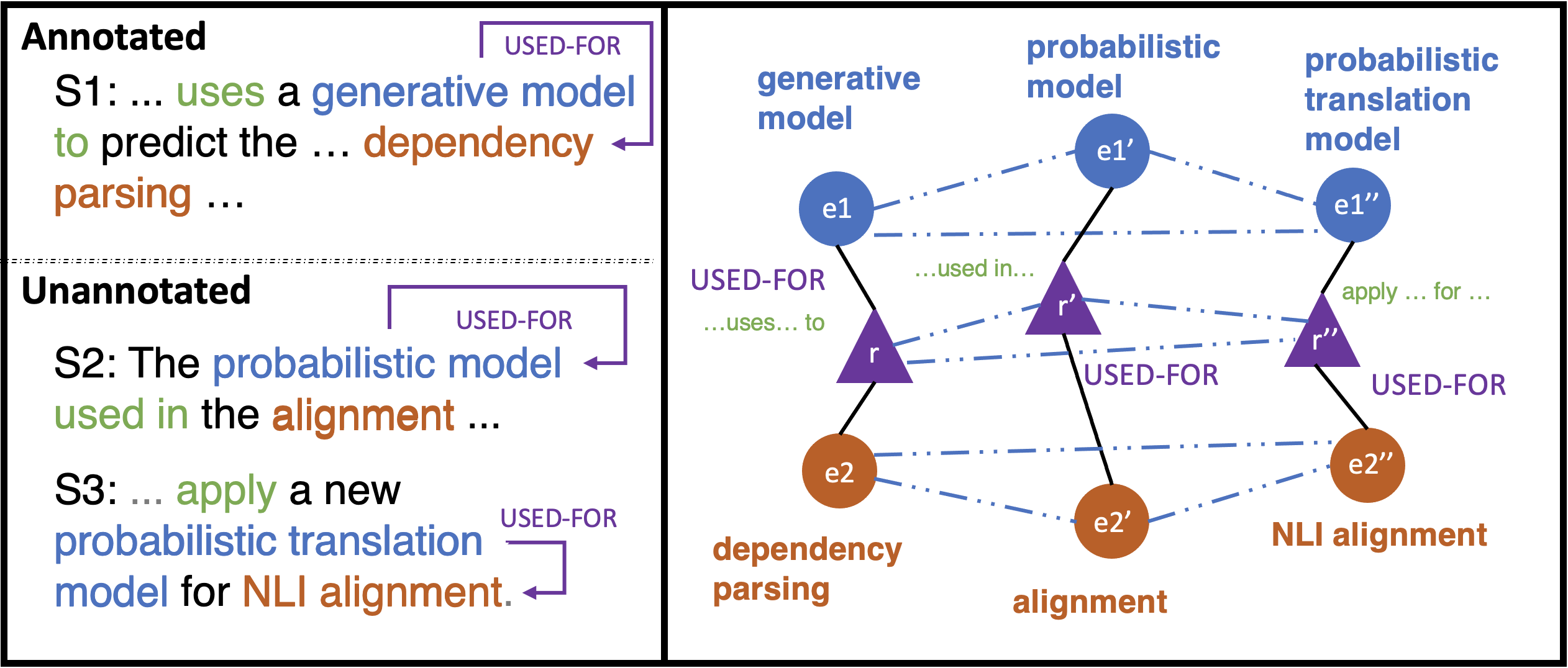}
 \label{fig:figure-1.b}
 }
 \caption{An example of label propagation. We represent the sentence as a triplet ($entity_1$, relation, $entity_2$) which consists of an entity pair (circle) and a relation (triangle) in a graph structure. The colored nodes indicate labeled semantic units (entity or relation candidates), while the uncolored nodes represent the unlabeled semantic units. Purple denotes relation label ${\texttt{Used-for}}$, blue denotes for entity label ${\texttt{Method}}$, and orange denotes another entity label label ${\texttt{Task}}$. 
 }
 \label{fig:intro1}
\end{figure}

SSL in NER and RE have performed very well in recent years by employing bootstrapping, distant supervision or graph-based approach  \citep{batista-etal-2015-semi, zeng-etal-2015-distant, pmlr-vR5-delalleau05a}. However, they either train a NER \citep{yang-etal-2018-distantly,chen-etal-2018-variational,lakshmi-narayan-etal-2019-exploration} or a RE model \citep{DualRE, hu-etal-2021-semi-supervised,9534434}. Therefore, they neglect the underlying connections between entity recognition and relation extraction under a semi-supervised learning scheme, making it harder for the model to assign the accurate annotation to unlabeled data. For instance, in Figure \ref{fig:intro1}, the annotated entity entity ``\textit{generative model}'' in sentence S1 and the unannotated ``\textit{probabilistic model}'' in sentence S2 are syntactically similar. Likewise, the context phrases ``\textit{uses... to}'' and ``\textit{used in}'' are also similar. If such similarities are ignored, the model may fail to draw a syntactic analogy between ``\textit{dependency parsing}'' and ``\textit{alignment}'', and thereby miss labeling the latter as an entity that shares the same type with the former. To the best of our knowledge, there is no universal framework to integrate the semi-supervised learning for different tasks in IE, despite evidence of the effectiveness of a joint or multi-task learning approach \citep{luan-etal-2018-multi, luan-etal-2019-general, DBLP:journals/corr/abs-2109-06067, luan-etal-2018-multi, luan-etal-2019-general, lin-etal-2020-joint}.

In addition, existing semi-supervised approaches devote considerable effort to aligning labeled and unlabeled data but do not exploit similarities between instance pairs that are structurally parallel, which exist across unlabeled data. Consequently, they do not perform classification from the perspective of global consistency. For example, given the sentences S1 to S3 in Figure \ref{fig:intro1}, we expect a system to recognize the entities and relations as (${\texttt{Method}}$, ${\texttt{Used-for}}$, ${\texttt{Task}}$) in triplet form. However, it is hard to infer the correct pseudo label to the unlabeled entities ``\textit{alignment}'' or ``\textit{NLI alignment}'' from the annotated entity ``\textit{dependency parsing}''. Because they are not semantically or lexically similar. Likewise, the affinity between ``\textit{uses to}'' and ``\textit{apply}'' is not obvious; and hence it would be difficult to extract the relation ${\texttt{Used-for}}$ between entities. Nonetheless, the ``\textit{alignment}'' and ``\textit{NLI alignment}'' pair are alike, and so are the pair ``\textit{probabilistic model}'' and ``\textit{probabilistic model}''. Exploiting the relationships between unlabeled data would integrate the information hidden in the text and make use of the large quantity of unlabeled data for semi-supervised learning. 

To address the above limitations, we propose a semi-supervised method based on label propagation over a heterogeneous candidate graph to populate labels for the two tasks (see Figure \ref{fig:core}). More specifically, we introduce a joint semi-supervised algorithm for the two tasks, where unannotated and annotated candidates (entities and relations) are treated as nodes in a heterogeneous graph, and labels are propagated across the graph through similarity-scored edges. Our framework {\modelname} considers the interactions among the unlabeled data by constructing the graph using the union of labeled and unlabeled data into one learning diagram. We evaluate {\modelname} on multiple benchmark datasets and our proposed framework achieve state-of-the-art results on both semi-supervised NER and RE tasks. To the best of our knowledge, this is the first work that performs semi-supervised learning for entity and relation extraction in a unified framework to leverage unannotated data for both tasks.

Our contributions are summarized as following:

\begin{itemize}
    \item  We propose a joint learning scheme using heterogeneous graph-based label propagation for semi-supervised NER and RE. The model exploits the interrelations between labeled and unlabeled data and the similarity between unlabeled examples from both tasks by propagating the information across a joint heterogeneous graph. To the best of our knowledge, this is the first work that combines semi-supervised NER and RE.
    
    \item We propose a unified semi-supervised framework for both entity and relation extraction. The framework generates candidate spans from the unlabeled data, automatically constructs a semantic similarity-based graph for all the candidates, and performs label propagation across the graph. 
    
    \item We show that our proposed method can reliably generate labels for unlabeled data and achieve good performance under a limited data scenario. Our model outperforms strong baselines in two- and single-task settings and establishes new state-of-the-art F1 on benchmark datasets.
\end{itemize}

\section{Related Work}



\paragraph{Joint Entity and Relation Extraction}  Name Entity Recognition, and Relation Extractions are two essential problems in information extraction \citep{3fc8a29e6f5c4b38be0fe8e5cc549d29}. Exploiting their interrelationships, models that combine the identification of entities and relations have attracted attention. Conventional joint extraction systems combine the tasks in a pipelined approach (e.g., \citet{ratinov-roth-2009-design, chan-roth-2011-exploiting,luu2014taxonomy,luu2015incorporating,tuan2016utilizing}): first identifying entities and employing the detected entity for relation extraction. However, they overlook their inherent correlation. Recent works have proposed coupling various IE tasks to avoid error propagation issues. For example, joint extract entities and relations \citep{miwa-sasaki-2014-modeling, li-ji-2014-incremental,luu2016learning} or end-to-end multi-task learning \citep{luan-etal-2018-multi, luan-etal-2019-general, Wadden2019EntityRA,lin-etal-2020-joint, zhang-etal-2017-end}. Despite evidence of the efficiency of joint or multi-task learning, there is currently no framework that integrates semi-supervised learning for both tasks in a joint entity and relation extraction system.

\paragraph{Semi-supervised learning} The Semi-Supervised learning seeks to enhance limited labeled data by leveraging vast volumes of unlabeled data \citep{semidef} which mitigate data-hungry bottleneck and supervision cost. SSL has a rich history \citep{1053799}. There have been substantial works in semi-supervised settings in NLP, such as bootstrapping \citep{gupta-manning-2014-improved, gupta-manning-2015-distributed, batista-etal-2015-semi}, co-training  \citep{10.1145/279943.279962}, distant supervision \citep{zeng-etal-2015-distant, yang-etal-2018-distantly}, and graph-based methods \citep{pmlr-vR5-delalleau05a, JMLR:v12:subramanya11a, 10.5555/1870658.1870675, luan-etal-2017-scientific}.

In particular, graph-based SSL algorithms have gained considerable attention \citep{Zhu2002LearningFL, labelandunlabel,pmlr-vR5-delalleau05a}. There are two underlying assumptions for the label propagation \citep{NIPS2003_87682805}. First, similar training samples are more likely to belong to the same class. Second, nodes in similar structures are likely to have the same label. Label propagation exploits these assumptions by propagating label information to surrounding nodes based on their proximity. The metric-based method had been applied in a graph-based SSL setting for its ability to infer labels for unseen classes directly during inference. For example, \citet{luan-etal-2017-scientific} propagates the label based on estimating the posterior probabilities of unlabeled data. Meanwhile, \citet{liu2019fewTPN} sought to exploit the manifold structure of novel class space in a transitive setting.

\section{Methodology}



\paragraph{Problem Definition}  The input of the problem is a sentence $X = \{x_1, ..., x_n\}$ consisting of n tokens, from which we derive $S = \{s_1,...,s_d\}$, the set of all possible within-sentence word sequence spans (up to length $L$) in the sentence. Let START$(i)$ and END$(i)$ denote the start and end indices of span $s_i$, $\mathcal{E}$ denote a set of predefined entity types, and $\mathcal{R}$ denote the set of relational types. The full data is defined as $D = (X,Y)$. In \modelname, the goal is to learn from the small portion of labelled data $D_l$ and generalize to the unlabelled portion of data $D_u$. The labelled data $D_l$ and unlabelled data $D_u$ are originally split from the training set $D_{train}$, where $D_l \cap D_u = \emptyset$. 

The purpose of this work is to predict a possible entity type $y_e(s_i) \in \mathcal{E}$ for each span $s_i \in S$ while predicting a possible relation types $y_r(s_i,s_j) \in R$ for every pair of spans $s_i \in S, s_j \in S$ under SSL settings. The  label can also be a `null' label for a span (i.e. $y_e(s_i) = \epsilon$) or a span pair (i.e. $y_r(s_i,s_j) = \epsilon$). The output of the task are $Y_e = \{ (s_i,e):si \in S,e \in \mathcal{E} \}$ and $Y_r = \{(s_i,s_j, r):s_i, s_j \in S, r \in \mathcal{R} \}$. 


\paragraph{Model Overview}

\begin{figure*}[t!]
\small
 \centering
 \includegraphics[width=0.9\textwidth]{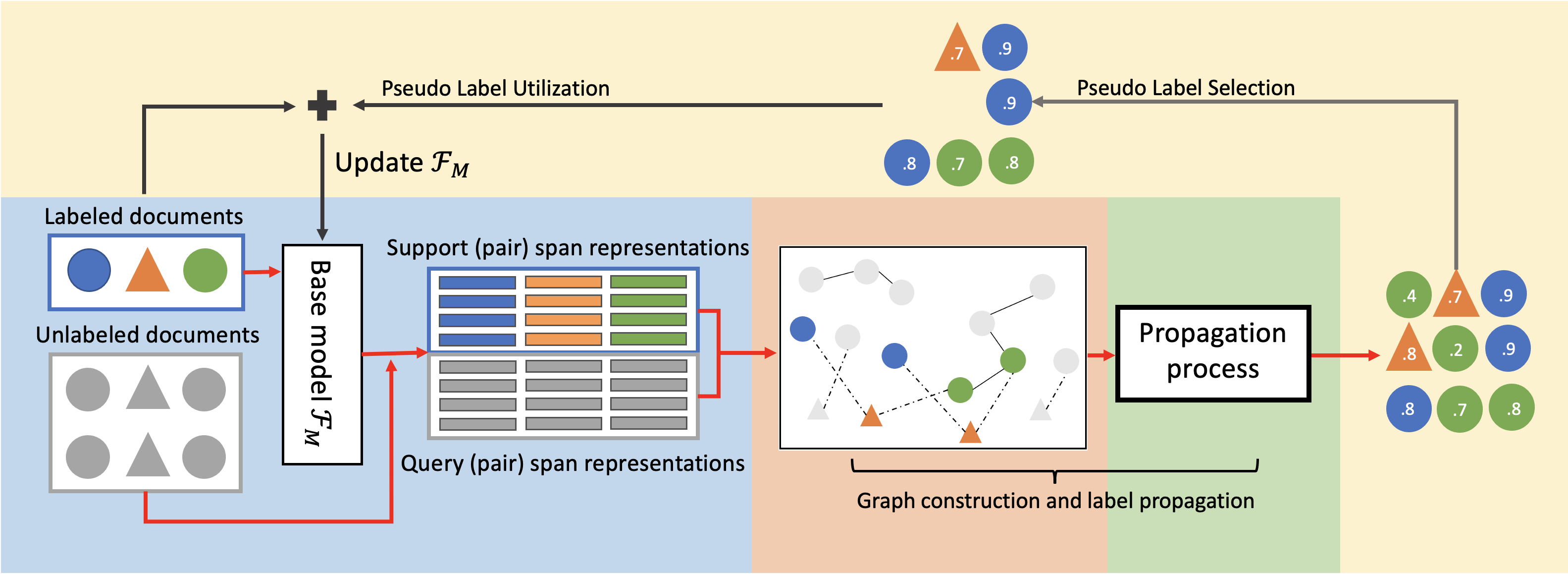}
 \caption{Overview of our proposed framework for semi-supervised joint learning. {\color{CornflowerBlue}\textsc{Span feature generation}}, {\color{orange}\textsc{Heterogeneous Graph Construction}}, {\color{YellowGreen}\textsc{Joint Label Propagation}} are represented in \textbf{{\color{red} red arrows}}, and {\color{Dandelion}\textsc{Model optimization}} is illustrated in \textbf{black arrows}}. 
 \label{fig:overall}
\end{figure*}

Figure \ref{fig:overall} illustrates an overview architecture of the proposed {\modelname} framework. Our framework consists of 1) \textsc{Span feature generation}  that learns the discriminative contextualized features for labelled data $D_l$ and unlabeled span $D_u$; 2) \textsc{Heterogeneous Graph Construction} which maps both labelled-unlabeled, labelled-labelled and unlabeled-unlabeled relationships for both entities and relations; 3) \textsc{Joint Label Propagation} which disseminates labels over the whole heterogeneous graph is produced by unlabeled nodes, and 4) \textsc{Model decode and fine-tune module} that decodes and select the refined propagated pseudo labels to perform fine-tuning. 

 \subsection{Span feature generation}
 Our feature extractor is a standard span-based model following prior work \citep{Wadden2019EntityRA, luan-etal-2018-multi, luan2018multitask}. For each input token $x_k$, we obtain contextualized representations $\mathbf{x}_k$ using a pre-trained language model (e.g., BERT \citep{devlin-etal-2019-bert}). For the i-th span $s_k \in S$, the span representation $\boldsymbol{h}_e(s_i)$ is as follows:
\begin{equation}
\boldsymbol{h}_e(s_i) = [\mathbf{x}_{START(i)};\mathbf{x}_{END(i)};\phi (s_i)] 
    \label{eq:ner-span-emb}
\end{equation}

\noindent where $\phi (s_i) \in \mathbb{R}^{1 \times d_{F}} $ denotes the learned embeddings of span width features.

For each pair of spans input $s_i, s_j \in S$, the span pair representation is defined as:
\begin{equation}
 \boldsymbol{h}_r(s_i, s_j) = [\boldsymbol{h}_e(s_i);\boldsymbol{h}_e(s_j);\mathcal{F}_{af}]
    \label{eq:re-span-emb}
\end{equation}

\noindent where $\mathcal{F}_{af} = \boldsymbol{h}_e(s_i) \cdot \boldsymbol{h}_e(s_j)$ refers to the entity affinity function of $e(s_i)$ and $e(s_j)$.  

 Both pairwise span feature $\boldsymbol{h}_r(s_i, s_j)$ and span feature $\boldsymbol{h}_e(s_i)$ will be fed into feedforward neural networks (FFNNs) respectively. The probability distribution of entity is denoted as $P_e(e|s_i), (e \in \mathcal{E} \cup \epsilon )$ and  entity pairs is denoted as $P_r(r|s_i, s_j), (r \in \mathcal{R}  \cup \epsilon)$.

The classification loss will be defined as:
\begin{equation}
    \mathcal{L} = \sum_{t \in T}w_t{\mathcal{L}_t}
\end{equation}

\noindent where $w_t$ is the predefined weight of a task $t$ and T is the total number of tasks.

We then use labelled data $D_l$ to train the classifier $C_l$. The $C_l$ can generate contextualized span or span pair feature from Equation~\ref{eq:ner-span-emb} and Equation~\ref{eq:re-span-emb} which converts unlabeled data $D_u$ into query entity presentation $\boldsymbol{h}_{u,e}$ or query entity pair representation $\boldsymbol{h}_{u,r}$. For labelled data $D_l$, we denote the $C_l$ generated support entity presentation as $\boldsymbol{h}_{l,e}$ and support entity pair representation as $\boldsymbol{h}_{l,r}$.

\subsection{Joint Semi-supervised Learning}

\paragraph{Heterogeneous Graph Construction} \label{the_alpha_k}

\begin{figure*}[ht!]
 \centering
 \includegraphics[width=0.8\textwidth]{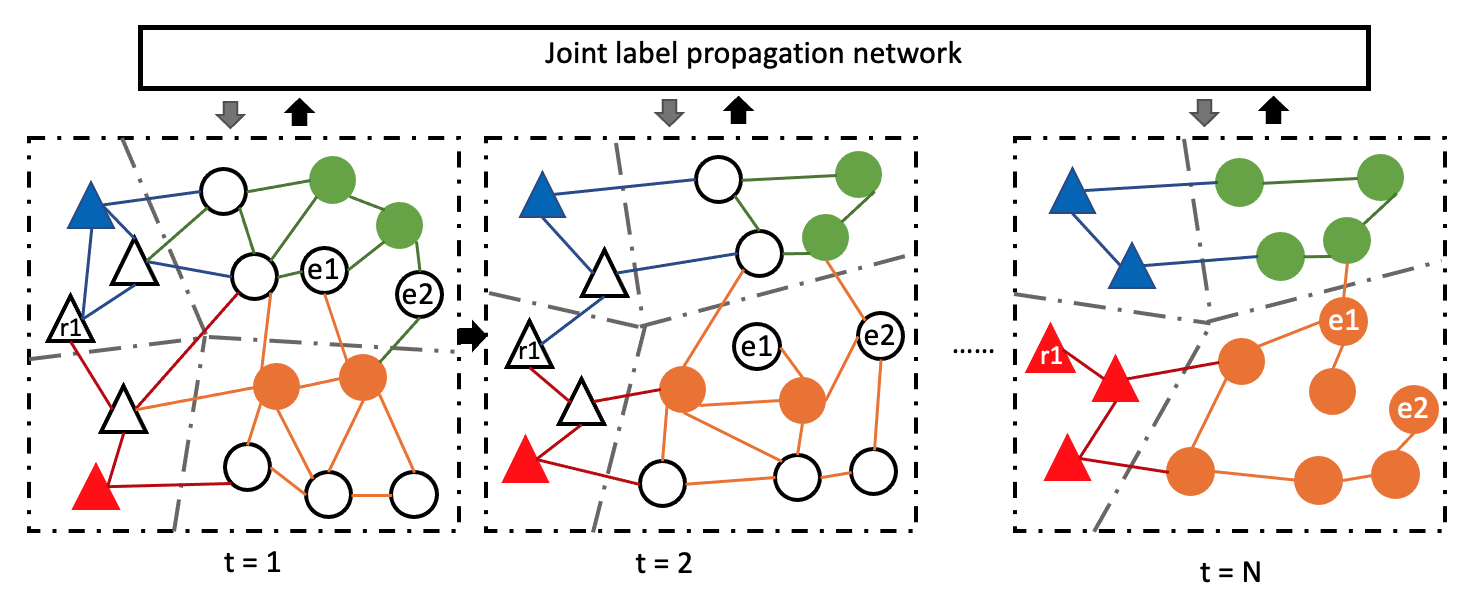}
 \caption{A conceptual demonstration of the label propagation process. Through the heterogeneous graph, our proposed joint semi-supervised learning method propagates labels to entity or relation candidates in the unlabeled data alternatively. As shown in the figure, the pseudo label for entities or relations will be refined every time t until stabilized.}
 \label{fig:core}
\end{figure*}

 We construct the heterogeneous graph to exploit the manifold structure of the class space and exploit the combination of labelled data $D_l$ and unlabeled data $D_u$. Specifically, we first examine the similarity relations among pairs of unlabeled query data as well as the similarity relationships between the labelled support data in order to take advantage of the smoothest constraints among neighbouring query unlabelled data in our semi-supervised joint entity and relation extraction task. 
 Based on the embedding space, we propose the use of transductive label propagation to construct a graph from the labelled support set and unlabeled query set, and then propogate the labels based on random walks to reason about relationships in labelled support and unlabeled query sets.
Figure \ref{fig:core} illustrates the whole process of heterogeneous graph-based propagation $\mathcal{G}$. The circle node is the entity span representation and the triangle node is the relation representation. 

For computational efficiency, we construct a k Nearest Neighbor (kNN) graph instead of a fully-connected graph. Let $N$ be the number of labelled entity representations and let $M$ be the number of unlabelled entity representations. Specifically, we take $N$ entity representations and $M$ unlabelled entity representations as nodes of an entity graph with size $T_e = N + M$. For the relation graph, we take span pair representation as nodes with size $T_r = ((N + M) \times (N + M))$. We construct a sparse affinity matrix, denoted as $\mathbf{A} \in \mathbb{R}^{\mathbf{T} \times \mathbf{T}}$, where $T_e, T_r \in \mathbf{T}$ by computing
the Gaussian similarity function between each node:
\begin{equation}
    \mathbf{A}_{ab} = exp(-\frac{||(\textbf{h}_a, \textbf{h}_b)||_{2}^2}{2 \sigma^2})
    \label{eq: weighted_matrix}
\end{equation}

\noindent where $\textbf{h}_a$ denotes the a-th entity representation or pairwise entity representation (i.e. $\{h_r(s_i, s_j), h_e(s_i), h_e(s_j)\} \in \textbf{h}_a$). The $\sigma$ is the length scale parameter.

Subsequently, we symmetrically normalize the non-negative and symmetric matrix $\mathbf{O} = \mathbf{A} + \mathbf{A}^T$ by applying Normalized Graph Laplacian on $O$:
\begin{equation}
    S = H^{(-1/2)}OH^{(-1/2)}
\end{equation}

\noindent where $H$ is a diagonal matrix with its ($i,i$)-value to be the sum of the i-th row of $O$.

For pairwise span representation $\boldsymbol{h}_r(s_i, s_j)$ is essentially a function of $\boldsymbol{h}_e(s_i)$ and $\boldsymbol{h}_e(s_ij)$. The entity nodes and the relation nodes are automatically associated via their representation.

\paragraph{Label propagation} We define a label matrix $Z \in \mathbb{R}^{V \times U}$ where U is either the size of entity types or relation types $ U = \{\mathcal{E};\mathcal{R}\}$. For label matrix, $Z$, the corresponding labelled data are one-hot ground truth labels and the rest are 0. Additionally, we denote $Y_t$ as a representation of the predicted label distributions at iteration t. Initially, we set the rows in $Y_0 = Z$. Starting from $Y_0$, message passing via label propagation in an iterative manner selects the type of the span or span pairs in the unlabeled set $D_u$ according to the graph structure according to the following operation:
\begin{equation}
    Y_{t+1} = c S Y_t + (1-c)Z
    \label{eq: hyper-parameter c}
\end{equation}

\noindent where $c \in (0,1)$ controls the probability of information being obtained from a node's adjacency nodes or its initial label. $Y_t$ refers to the predicted labels at time $t$. 

Given $Y_0 = Z$, and equation \eqref{eq: hyper-parameter c} , we have: 
\begin{equation}
    Y_{t} = (cS)^(t-1)Z + (1-c) \sum^{t-1}_{i=0} (cS)^i Y
    \label{eq: yt}
\end{equation}

As the parameter $c \in (0,1)$, taking the limit of equation \eqref{eq: yt}~$(t->\infty)$ we have: 
\begin{equation}
 \lim_{t\to\infty} Y_{t} = (1-c)(1-cS)^{-1} = Y_{converge}
\end{equation}

The label propagation will converge to $Y_{converge}$.

  



\subsection{Model optimization}

After we obtain the $Y_{converge}$, we use the $softmax$ function followed by a standard \verb|argmax| operation to determine the pseudo labels $\{\hat y\}$ for all the instances in the unlabeled set based on the final label probability matrix $Y_{converge}$. After generating the pseudo labels $\{\hat y\}$ for all the labelled data $D_l$, we filter those of lower quality with a confidence threshold of $g$ and combine the rest (of confidence above the threshold) with the labelled data $D_l$ to retrain the classification model:
\begin{gather*}
    \{\hat y\} = \{\hat y \text{ | confidence}(y) \geq g\}    \\
    (X, Y) = (X,Y)_{D_l} + \{(x,\hat y)|x \in D_u\}
\end{gather*}
\indent As shown in the Figure~\ref{fig:overall}, the final step in our proposed joint semi-supervised learning framework is re-training. The retraining model is exactly the same as the baseline model and the joint NER-RE objectives also remain the same

\section{Experiments}


We evaluate the effectiveness of {\modelname} against models from two lines of work: semi-supervsied NER and semi-supervsied RE. We also provided a detailed analysis to demonstrate the benefits of our framework. For implementation details and dataset descriptions please refer to Appendix~\ref{Implement_D} and Appendix~\ref{datesets}.

\paragraph{Datasets} We perform experiments to assess the efficacy of our framework on four public datasets: SciERC \citep{luan2018multitask}, ACE05 \citep{ACE05}, SemEval \citep{hendrickx2019SemEval2010} and ConLL \citep{DBLP:journals/corr/cs-CL-0306050}.

\begin{table*}[t!]
\centering

\resizebox{\textwidth}{!}{%
\begin{tabular}{cccccccccccccc}
\hline
\multirow{2}{*}{\textbf{Settings\% labeled Data}} & 
\multirow{2}{*}{\textbf{Task}} & \multicolumn{3}{c}{\textbf{5\%}} & \multicolumn{3}{c}{\textbf{10\%}} & \multicolumn{3}{c}{\textbf{20\%}} & \multicolumn{3}{c}{\textbf{30\%}}  \\ 
\cmidrule(lr){3-5} 
\cmidrule(lr){6-8} 
\cmidrule(lr){9-11} 
\cmidrule(lr){12-14}
 & & P & R & F1 & P & R & F1 & P & R & F1 & P & R & F1  \\ \hline
\multirow{2}{*}{{\Large\basemodelname}(baseline)}
& NER & 
46.78 & 47.25 & 47.01 &
52.44 & 59.80 & 55.94 &
55.80 & 62.37 & 58.90 & 
60.42 & 67.56 & 63.79   \\
 & RE & 
 20.89 & 15.40 & 17.73 & 
 35.75 & 16.74 & 22.80 & 
 38.68 & 23.51 & 29.25 & 
 43.41 & 29.77 & 35.32   \\\hline
\multirow{2}{*}{{\Large\modelname}}  
 & NER &  
 \textbf{52.67} & \textbf{48.46} & \textbf{51.02} &  
 \textbf{60.15} & \textbf{61.95} & \textbf{61.04} &  
 \textbf{62.03} & \textbf{64.52} & \textbf{63.25} &  
 \textbf{66.55} & \textbf{65.73} & \textbf{66.19}    \\
 & RE &  
 \textbf{40.82} & \textbf{33.78} & \textbf{36.97} &  
 \textbf{44.42} & \textbf{26.34} & \textbf{39.98} &
 \textbf{44.55} & \textbf{45.28} & \textbf{44.91} &  
 \textbf{57.94} & \textbf{39.32} & \textbf{46.85} \\ \hline

\end{tabular}%
}
\caption{Performance on SciERC with various amount of labeled data.}
\label{tab: scierc result}
\end{table*}

\begin{table*}[t!]
\centering

\resizebox{\textwidth}{!}{%
\begin{tabular}{cccccccccccccc}
\hline
\multirow{2}{*}{\textbf{Settings\% labeled Data}} & 
\multirow{2}{*}{\textbf{Task}} & \multicolumn{3}{c}{\textbf{5\%}} & \multicolumn{3}{c}{\textbf{10\%}} & \multicolumn{3}{c}{\textbf{20\%}} & \multicolumn{3}{c}{\textbf{30\%}}  \\ 
\cmidrule(lr){3-5} 
\cmidrule(lr){6-8} 
\cmidrule(lr){9-11} 
\cmidrule(lr){12-14}
 & & P & R & F1 & P & R & F1 & P & R & F1 & P & R & F1  \\ \hline
\multirow{2}{*}{{\Large\basemodelname} (baseline)}
& NER & 
78.32 & 76.88 & 77.59 &
80.81 & 81.68 & 81.24 &
81.01 & 85.17 & 83.04 & 
84.51 & 86.98 & 85.72   \\
 & RE & 
 46.33 & 20.85 & 28.76 & 
 49.10 & 30.76 & 37.82 & 
 46.71 & 46.31 & 46.51 & 
 57.59 & 48.78 & 52.83   \\\hline
\multirow{2}{*}{{\Large\modelname}}  
 & NER &  
 \textbf{81.91} & \textbf{78.39} & \textbf{80.11} &  
 \textbf{83.38} & \textbf{82.76} & \textbf{83.07} &  
 \textbf{86.82} & \textbf{83.76} & \textbf{85.27} &  
 \textbf{87.69} & \textbf{86.40} & \textbf{87.04}    \\
 & RE &  
 \textbf{48.10} & \textbf{29.63} & \textbf{36.67} &  
 \textbf{48.89} & \textbf{36.23} & \textbf{42.00} &
 \textbf{60.26} & \textbf{44.67} & \textbf{51.30} &  
 \textbf{61.54} & \textbf{48.65} & \textbf{54.34} \\ \hline

\end{tabular}%
}
\caption{Performance on ACE05 with various amounts of labelled data.}
\label{tab: ace05 result}
\end{table*}

\begin{table*}[t!]
\small
\centering
\resizebox{\textwidth}{!}{%
\begin{tabular}{cccccccccc}
\hline
\multirow{2}{*}{\textbf{Methods / \% labeled Data}} & \multicolumn{3}{c}{\textbf{5\%}} & \multicolumn{3}{c}{\textbf{10\%}} & \multicolumn{3}{c}{\textbf{30\%}} 
\\\cmidrule(lr){2-4} 
\cmidrule(lr){5-7} 
\cmidrule(lr){8-10}
 & P & R & F1 & P & R & F1 & P & R & F1 \\ \hline
 
Mean-Teacher
& 70.33 & 68.55 & 69.05 
& 74.01 & 72.08 & 73.37
& 79.09 & 82.23 & 80.61 \\ 
Self-Training 
& 73.10 & 70.01 & 71.34 
& 75.54 & 73.00 & 74.25
& 80.92 & 82.39 & 81.71 \\ 
DualRE 
& 73.32 & 77.01 & 74.35 
& 75.51 & 78.81 & 77.13 
& 81.30 & 84.55 & 82.88 \\ 
MRefG 
& 73.04 & 78.29 & 75.48
& 76.32 & 79.76 & 77.96 
& 81.75 & 84.91 & 83.24 \\ 
MetaSRE 
& 75.59 & 81.40 & 78.33 
& 78.05 & 82.29 & 80.09
& 82.01 & 87.95 & 84.81 \\ 
GradLRE 
& 75.96 & 83.72 & 79.65 
& 78.90 & 82.94 & 81.69 
& 82.74 & 88.49 & 85.52 \\ \hline


{\modelname}$\dagger$ 
& \textbf{76.09} & \textbf{86.35} & \textbf{80.89}  
& \textbf{79.10} & \textbf{88.64} & \textbf{83.60}   
& \textbf{83.62} & \textbf{89.35} & \textbf{86.39} \\ \hline

Gold labels
& - & - & 84.64
& - & - & 84.40
& - & - & 87.08 \\ \hline
\end{tabular}%
}
\caption{Performance on SemEval with various labelled data and 50\% unlabeled data. We provide the \textit{Gold labels} serves as the upper bound of the model. ($\dagger$ indicates our framework.)}
\label{tab:SemEval_performance}
\end{table*}

\begin{table*}[t!]
\small
\centering
\resizebox{\textwidth}{!}{%
\begin{tabular}{cccccccccc}
\hline
\multirow{2}{*}{\textbf{Methods / \% labeled Data}} & \multicolumn{3}{c}{\textbf{5\%}} & 
\multicolumn{3}{c}{\textbf{10\%}} & 
\multicolumn{3}{c}{\textbf{30\%}} \\
\cmidrule(lr){2-4} 
\cmidrule(lr){5-7} 
\cmidrule(lr){8-10}
 & P & R & F1 & P & R & F1 & P & R & F1 \\ \hline
VSL-GG-Hier & 84.13 & 82.64 & 83.38 &
84.90 & 84.52 & 84.71 &
85.37 & 85.67 & 85.52 \\ 
MT + Noise & 83.74 & 81.49 & 82.60 &
 84.32 & 82.64 & 83.47 &
 84.98 & 84.78 & 84.88 \\ 
Semi-LADA & 86.93 & 85.74 & 86.33 & 
88.61 & 88.95 & 88.78 & 
89.98 & 90.52 & 90.25 \\ \hline
{\modelname}$\dagger$   & 
 \textbf{89.88} & \textbf{85.98} & \textbf{87.68} &
 \textbf{88.76} & \textbf{90.25} & \textbf{88.89} &
 \textbf{91.16} & \textbf{90.58} & \textbf{90.87} \\\hline

\end{tabular}%
}
\caption{Performance on CoNLL 2003 with various labelled data. ($\dagger$ indicates our framework.)}
\label{tab:CoNLL_performance}
\end{table*}

\subsection{Main Results}
Tables~\ref{tab: scierc result} to~\ref{tab:CoNLL_performance} provide the framework performance on the joint entity and relation extraction task, the NER task, and the RE task, respectively. Note that {\basemodelname} only trains using the labelled corpus. (i.e., The {\basemodelname} only trains with $5\%$, $10\%$ and $30\%$ training data.) As no unlabeled data are used in the training, this indicates the lower bound performance and establishes a new baseline. 

\paragraph{Results on SciERC} Table~\ref{tab: scierc result} illustrate our main results on semi-supervised joint learning on the SciERC dataset. We observed {\modelname} improve significantly on both entity recognition and relation extraction. {\modelname} achieves 3.97\% and 15.89\% F1 improvements, respectively, comparing to {\basemodelname}. This improvement validates the robustness of {\modelname} by performing joint learning on NER and RE.

\paragraph{Results on ACE05} Table~\ref{tab: ace05 result} we summarize the results of comparing to the baseline performance. As can be seen from the table, {\modelname} improves by around 2\% and 5\% on F1 for entity recognition and relation extraction task respectively. The results of this study provide further evidence of the consistency of the framework for multitask datasets.

\paragraph{Results on SemEval} Table \ref{tab:SemEval_performance} summarizes the experimental results on the SemEval dataset using various labelled data and 50\% unlabeled data. {\modelname} improves on the {\basemodelname} by 5.47\% on average. We can observe that {\modelname} attains 1.24\%, 1.91\% and 0.81\% F1 improvements over the state-of-the-art model GradLRE \citep{hu2021gradient} with 5\%, 10\% and 30\% training data. Moreover, the model's performance consistently improves while narrowing down the gap towards the upper bound as the proportion of labelled data increases. {\modelname} establishes a new state-of-the-art result, indicating that our framework is relatively robust even when performing a single task: semi-supervised RE.

\paragraph{Results on CoNLL}

Experimental results on CoNLL dataset are shown in Table~\ref{tab:CoNLL_performance}. Semi-LADA \citep{chen-etal-2020-local} is the current state-of-the-art semi-supervised NER model. In multiple training data settings, {\modelname} achieves an average improvement of 0.9\% over Semi-LADA. Semi-LADA reports a 91.83\% F1 score in a fully supervised setting, as the upper bound of the semi-supervised model. {\modelname} achieves 90.87\% in F1 score with 30\% of training data. The difference between the upper bound and the model performance narrows to less than 1\%. Moreover, {\modelname} surpasses the current state-of-the-art semi-supervised NER model, showing our model's effectiveness on another single task: semi-supervised NER.

\subsection{Analysis}


\begin{table*}[th!]

\centering
\resizebox{\textwidth}{!}{
\begin{tabular}{ccccccccccccc}
\hline
\multirow{2}{*} & \multicolumn{12}{c}{\textbf{Name Entity Recognition}} \\ \cline{2-13} 
 {\textbf{Model / Task}} & 
 \multicolumn{3}{c}{\textbf{5\%}} & \multicolumn{3}{c}{\textbf{10\%}} & \multicolumn{3}{c}{\textbf{20\%}} & \multicolumn{3}{c}{\textbf{30\%}} \\ 
 \cmidrule(lr){2-4} 
\cmidrule(lr){5-7} 
\cmidrule(lr){8-10} 
\cmidrule(lr){11-13}
 & P & R & F1 & P & R & F1 & P & R & F1 & P & R & F1 \\ \hline

{\basemodelname}
 & 46.78 & 47.25 & 47.01  
 & 52.44 & 59.80 & 55.94  
 & 55.80 & 62.37 & 58.90  
 & 60.42 & 67.56 & 63.79  \\

{\woRE}
&  51.82 & 45.10 & 48.23
&  58.92 & 53.46 & 56.06 
&  61.55 & 57.77 & 59.60
&  64.71 & 63.32 & 64.01\\
{\modelname}
 & \textbf{52.67} & \textbf{48.46} & \textbf{51.02}  
 & \textbf{60.15} & \textbf{61.95} & \textbf{61.04}  
 & \textbf{62.03} & \textbf{64.52} & \textbf{63.25} 
 & \textbf{66.55} & \textbf{65.73} & \textbf{66.19} \\ \hline

\end{tabular}
}
\caption{Ablation study on pure NER task on SciERC dataset.}
\label{tab:ablation_ner}
\end{table*}

\begin{table*}[th!]
\centering
\resizebox{\textwidth}{!}{
\begin{tabular}{ccccccccccccc}
\hline
\multirow{2}{*} & \multicolumn{12}{c}{\textbf{Relation Extraction}} \\ \cmidrule(lr){2-4} 
\cmidrule(lr){5-7} 
\cmidrule(lr){8-10} 
\cmidrule(lr){11-13}
{\textbf{Model / Task}} & \multicolumn{3}{c}{\textbf{5\%}} & \multicolumn{3}{c}{\textbf{10\%}} & \multicolumn{3}{c}{\textbf{20\%}} & \multicolumn{3}{c}{\textbf{30\%}} \\ \cline{2-13} 
 & P & R & F1 & P & R & F1 & P & R & F1 & P & R & F1 \\ \hline
 
  {\basemodelname} 
 & 20.89 & 15.40 & 17.73 
 & 35.75 & 16.74 & 22.80 
 & 38.68 & 23.51 & 29.25  
 & 43.41 & 29.77 & 35.32  \\

{\woNER}
& 38.92 & 13.35 & 19.88 
& 19.20 & 44.97 & 26.91 
& 22.27 & 62.53 & 32.74 
& 32.12 & 44.56 & 37.33  \\

{\modelname}  
 & \textbf{40.82} & \textbf{33.78} & \textbf{36.97}  
 & \textbf{44.42} & \textbf{26.34} & \textbf{39.98}  
 & \textbf{44.55} & \textbf{45.28} & \textbf{44.91}  
 & \textbf{57.94} & \textbf{39.32} & \textbf{46.85} \\\hline

\end{tabular}
}
\caption{Ablation study on pure RE task on SciERC dataset.}
\label{tab:ablation_re}
\end{table*}

\begin{table*}[t!]
\centering
\small
\resizebox{\textwidth}{!}{
\begin{tabular}{p{0.4\textwidth}p{0.17\textwidth}p{0.17\textwidth}p{0.17\textwidth}} \hline
\textbf{Sentence} & \textbf{Semi-LADA} & \textbf{GradLRE} & \textbf{\modelname}  \\\hline

S1: We propose a {\color{red}Cooperative Model} for natural language understanding in a {\color{blue}dialogue system}.
& e1: Method \newline e2: Task  \newline R: -
& e1: - \newline e2: - \newline R: Used-for
& e1: Method \newline e2: Task \newline R: Used-for\\\hline

S2: We address appropriate {\color{red}user modelling} in order to generate cooperative responses to each user in {\color{blue}spoken dialogue systems}.  & e1: Method \newline e2: Task  \newline R: -
& e1: - \newline e2: - \newline R: Used-for (x)  
& e1: Method \newline e2: Task \newline R: Part-of \\\hline

S3: We explore {\color{red}correlation of dependency relation paths} to rank candidate answers in {\color{blue}answer extraction}.
& e1: - (x) \newline e2: Task  \newline R: -
& e1: - \newline e2: - \newline R: no\_relation (x)  
& e1: OST \newline e2: Task \newline R: Used-for \\\hline

S4: We present a {\color{blue}syntax-based constraint} for word alignment, known as the {\color{red}cohesion constrain}. 
& e1: Generic (x) \newline e2: Generic (x) \newline R: -
& e1: - \newline e2: - \newline R: Hyponym-of 
& e1: OST \newline e2: OST \newline R: Hyponym-of \\\hline

\end{tabular}%
}

\caption{Case study of {\modelname}. The red marked span denotes the head ($e1$) entity while the blue marked span represents the tail ($e2$) entity. Semi-LADA performs \texttt{OtherScientificTerm} abbreviated as \texttt{OST}. (x) indicates the wrong prediction and - means the model does not have certain predictions (i.e., The model does not predict entity type or relation type).}
\label{tab:ablation_re}
\end{table*}

\subsubsection{Ablation Studies}
This section provides comprehensive ablation studies to show the efficacy of {\modelname} frameworks. Tables~\ref{tab:ablation_ner} and ~\ref{tab:ablation_re} show the effect of joint label propagation on single-task (NER or RE) prediction accuracy. {\woRE} denotes ablating the relation propagation while {\woNER} denotes ablating the entity propagation. As a lower bound to the framework, we provide the {\basemodelname} result, which is the base model without any propagation.
As shown in Table~\ref{tab:ablation_ner}, although {\woRE} achieved an average 0.85\% improvement on F1 compared to {\basemodelname}. The {\modelname} further improve the performance significantly by 4.01\%, 4.98\%, 3.65\% and 2.19\% across 5\%, 10\%, 20\% and 30\% training data, respectively. 
From Table~\ref{tab:ablation_re}, we observed that {\woRE} attain an average of 2.94\% performance gain in F1 compared to {\modelname}. Though {\woRE} shows its effectiveness, {\woNER} has 7.03\% further overall across different proportions of training data. In general, we observe that joint label propagation is very helpful to {\modelname} performance, especially for relation extraction tasks.

We investigate a real and illustrative example in Figure \ref{fig:intro1}. Given sentences S1 to S3. {\woRE} is unable to identify the label of "alignment" in S2 and "NLI alignment" in S3. Moreover, {\woNER} tends to miss predict the pseudo label as \texttt{no\_relation}. More specifically, in annotated S1, the entity "dependency parsing" has no direct link to the entity "alignment" in S2 and entity "NLI alignment" in S3. Consequently, {\woRE} makes the wrong prediction. Similar to {\woRE}, the relation indicator "uses..to" in annotated S1 is semantically similar to "used in" in S2 but not akin to "apply..for.." in S3, hence {\woNER} miss identify the label of $r'''$. Whereas {\modelname} can assign the correct pseudo label to entities and relations in all three sentences for it benefits from the shared information from NER and RE. The results indicate that our framework {\modelname} could leverage the interactions across the two tasks and derive useful information from a broader context. Therefore achieve significant improvement across NER and RE tasks.

\subsubsection{Case study}

We perform a case study examining our framework's performance on four sentences (i.e., S1, S2, S3, and S4) in comparison to the benchmark models Semi-LADA and GradLRE. Semi-LADA performs semi-supervised NER task while GradLRE performs semi-supervised RE task. Meanwhile {\modelname} performs the semi-supervised style joint for NER and RE.

S1 has a simple structure, and all three models correctly classify the label for relation and entity. For S2, the GradLRE misclassifies the "Statistical machine translation" entity as Task. Most of the labelled samples with given entity pair are likely as in (e1: \texttt{Method}, e2: \texttt{Task}), plus there is a relation indicator "in order to," which misguides the GradLRE into the wrong prediction. Similarly, in S4, Semi-LADA predicts the entity as \texttt{Generic}, the dominant class in the training set. {\modelname} can assign the correct label without being sensitive to the label distribution in the training data.

Moreover, Semi-LADA fails to recognize the entity "correlation of dependency relation paths" in S3, while GradLRE cannot identify the relation \texttt{Used-for}. One possible reason is that there were not many similar long sequences in the training data. Consequently, Semi-LADA is insufficient in entity learning, especially for long lines, while the GradLRE fails to establish edges with samples in the training set. {\modelname} not only builds a connection between labelled and unlabeled data but also within labelled/unlabeled data. The extra connections hence help our model to make the correct prediction. 

\subsubsection{Qualitative Analysis} 

Table \ref{tab: prop performance} shows the qualitative results of our proposed method Joint Semi-supervised Learning for Entity and Relation Extraction with Heterogeneous Graph-based Propagation. We show the performance of the propagated pseudo labels with the ground truths under 10\% split training set on ACE05 dataset. As we can see from the performance Table \ref{tab: prop performance}, in both NER and RE, the recall of the predictions indicates that most of the positive candidates have been propagated a positive label.Meanwhile, the precision of the predictions for the NER task is also high. However, the precision for the RE task is low, showing that almost half of the null candidates have been assigned a positive label. The propagation of RE tasks is still quite challenging. 

\begin{table}[th!]
\centering
\small

\begin{tabular}{cccc}
\hline
   \% & P & R & F1   \\\hline
NER & 86.23 & 92.78 & 89.34 \\
RE  & 52.17 & 98.82 & 68.57 \\\hline
\end{tabular}
\caption{Qualitative results of our method in 10\% split on ACE05 dataset. (Average F1)}
\label{tab: prop performance}
\end{table}
In spite of this, our method still generally produces more accurate predictions. Given a sentence in ACE05: `Although the Russian government...'. Our model prediction for the phrase "Russian government" is "Organization", which is more accurate than the ground truth GPE-Geographic Entities.

\section{Conclusion}

In this paper, we propose a novel heterogeneous graph-based propagation mechanism for joint semi-supervised learning of entity and relation extraction. For the first time, we explore the interrelation between different tasks in a semi-supervised learning setting. We show that the joint semi-supervised learning of two tasks benefits from their codependency and validates the importance of utilizing the shared information between unlabeled data. Our experiments show that combining the two tasks boosts system performance. Moreover, by leveraging the inherent We also evaluate three public datasets over competitive baselines and achieve state-of-the-art performance. We also conduct ablation studies of our proposed framework, which demonstrate the effectiveness of our model. We further present case studies of our system output.

\section{Limitations}

\paragraph{May extend to other domains} In this paper, we present a generic framework and evaluate the effectiveness of our proposed model {\modelname} on three public datasets. We may further extend the framework to various datasets in different domains. For example, ACE05 \citep{ACE05} in social networks, journalism, and broadcasting, as well as GENIA corpus \citep{10.5555/1289189.1289260} in biomedical research.

\paragraph{May extend to other NLP tasks} Our proposed model focus on two tasks, namely NER and RE. We may extend our framework to include more information extraction tasks, such as coreference resolution and event extraction. Moreover, we may contract knowledge graphs from extracted structural information.

\section*{Acknowledgment}
This research is supported by Nanyang Technological University, under SUG Grant (020724-00001)

\bibliography{custom}
\bibliographystyle{acl_natbib}

\appendix


\section{Experimental Settings}\label{Implement_D}

\paragraph{Framework} We show our overall framework {\modelname} in Figure \ref{fig:overall}. Following \citep{liu2019fewTPN}, the hyper-parameter $c$ in Equation \ref{eq: hyper-parameter c} is set to 0.99. According to our empirical findings, the best values for the settings of k and $\sigma$ in graph construction in Section \ref{the_alpha_k} are varied in datasets. We select $\sigma$ as two and the k as 50. Meanwhile, We adopt the affinity function $\mathcal{F}_{af}$ with all the generated spans between relation spans. Moreover, we perform average pooling for them. The optimal hyperparameters and settings are selected based on the model's performance. 

\paragraph{Training}

We employ the BERT-cased as an encoder for SemEval and ConLL datasets and adopt the SciBERT-SCIVOCAB-cased \citep{beltagy-etal-2019-scibert} encoder for the SciERC dataset as suggested in \citep{luan-etal-2019-general}.  The rest will be treated as an unlabeled set. To maximize the loss, we use BERTAdam with a 1e-3 learning rate. The maximum span width is set at 8. 

\section{Datasets and baselines}\label{datesets}


\subsection{Dataset implementation} 

For semi-supervised joint task, we consider SciERC \citep{luan2018multitask} and ACE05\citep{ACE05} datasets and follow the pre-processing steps in \citep{Wadden2019EntityRA}. For a single task, we conduct experiments against models from two types of work: semi-supervised NER, and semi-supervised RE. For the semi-supervised NER task, we consider ConLL 2003 (ConLL) \citep{DBLP:journals/corr/cs-CL-0306050} and adopt the pre-processing in \citep{chen-etal-2020-local}. For semi-supervised RE we evaluate our approach on SemEval 2010 Task 8 (SemEval) \citep{hendrickx2019SemEval2010} dataset and adopt the pre-processing in \citep{hu2021gradient}. Note that the entity mentioned in the sentences in the \textbf{SemEval} has been identified and marked in advance.

Table \ref{tab: dataset statistic} shows the statistics of each dataset. 

\begin{table}[th!]
\centering
\small
\begin{tabular}{ccccccccc}
\hline
\multirow{2}{*}{\textbf{Dataset}} & \multicolumn{3}{c}{\textbf{Sentences}}  & \multicolumn{2}{c}{\textbf{Types}} \\ 
\cmidrule(lr){2-4} 
\cmidrule(lr){5-6} 
 & Train & Dev & Test & \# E & \# R \\ \hline
SciERC & 1861 & 275 & 551 & 6 & 7 \\
ACE05 & 10051 & 2424 & 2050 & 7 & 6 \\
ConLL & 14,987 & 3466 & 3684 & 4 & - \\
SemEval & 7199 & 800 & 1864 & - & 19 \\ \hline
\end{tabular}
\caption{\textbf{Statistics for the SciERC, ACE05, ConLL and SemEval datasets}. \#E: Number of entity classes. \#R: Number of relation classes.}
\label{tab: dataset statistic}
\end{table}

\paragraph{Data split for semi-supervised settings} We follow split settings in \citep{chen-etal-2020-local}, \citep{hendrickx2019SemEval2010} respectively for ConLL and SemEval and generate different proportions (5\%, 10\% and 30\%) of training data to investigate how training set size impacts performance and to retain the original development set and test set for evaluation purposes. Noted that we sample 50\% of the training set as the unlabeled set as \citep{hendrickx2019SemEval2010} for fair comparisons. For ACE05 and SciERC datasets, we split the training data based on documents and generate 5\%, 10\%l 20\% and 30\% of training data. In particular, we endeavour to ensure that each proportion of data contains as many types of entity types and relation types possible.

\subsection{Evaluation Metrics}
We consider the same criteria to apply as previous works \citep{hu2021gradient,hu-etal-2021-semi-supervised,9534434,DualRE} where precision and recall serve as supplementary metrics, while F1 score serves as the primary evaluation metrics. Note that the evaluation excludes the accurate prediction for \texttt{no\_relation}.

\subsection{Compared baselines}

\paragraph{Semi-supervised joint learning} For joint learning, because there is no prior study on semi-supervise joint learning, we use DYGIE++ \citep{Wadden2019EntityRA} (i.e. {\basemodelname}) as our baseline model to train.  

\paragraph{Semi-supervised NER}
In order to show that our {\modelname} framework works with unlabeled data, we compared it to three recent state-of-the-art semi-supervised NER models that were already in use: 
\begin{itemize}
    \item \textbf{VSL-GG-Hier} \citep{chen-etal-2018-variational} introduced a hierarchical latent variables models into semi-supervised NER learning.
    \item \textbf{MT + Noise} \citep{lakshmi-narayan-etal-2019-exploration} explored different noise strategies including word-dropout, synonym-replace, Gaussian noise and network-dropout in a mean-teacher framework.
    \item \textbf{Semi-LADA} \citep{chen-etal-2020-local} proposes a local additivity based data augmentation method which uses the back-translation technique.
\end{itemize}

\paragraph{Semi-supervised RE}
We compared our {\modelname} framework with the following 6 representative semi-supervised relation models:
\begin{itemize}

\item{\textbf{Mean-Teacher}}  promotes the model's variants to generate consistent predictions for comparable inputs.

\item{\textbf{DualRE}} \citep{DualRE} trains a prediction and retrieval module in conjunction to choose samples from unlabeled data.

\item{\textbf{MRefG}} \citep{9534434} constructs reference graphs to semantically relate unlabeled data to labelled data.

\item{\textbf{MetaSRE}} \citep{hu-etal-2021-semi-supervised} constructs pseudo labels on unlabeled data using meta-learning from the successfulness of the classifier module as an extra meta-objective. 

\item{\textbf{GradLRE}} \citep{hu2021gradient} is the state-of-the-art approach that encourages pseudo-labeled data to mimic the gradient descent direction on labelled data and boost its optimization capabilities via trial and error.

\item{\textbf{Gold labels}} train annotated (i.e., sampled 5\%, 10\% or 30\% training data) and unlabeled data with their gold labels indicating the model upper bound. 
    
\end{itemize}




\end{document}